\documentclass[11pt]{article}

\usepackage[utf8]{inputenc}
\usepackage[T1]{fontenc}
\usepackage{amsmath}
\usepackage{amssymb}
\usepackage{booktabs}
\usepackage{graphicx}
\usepackage{hyperref}
\usepackage{xurl}
\usepackage{microtype}
\usepackage{enumitem}
\usepackage{multirow}

\IfFileExists{neurips_2025.sty}{
  \usepackage[final,main]{neurips_2025}
}{
  \usepackage{geometry}
  \geometry{margin=1in}
}
\makeatletter
\renewcommand{\@notice}{}
\makeatother

\title{Associative-State Universal Transformers:\\
\large Pilot Studies in Structured Recurrent State, Sparse Retrieval, and Latent Compression}
\author{
Liu Xiao \\
\texttt{liu.xiao.in@gmail.com}
}
\date{}

\begin{document}
\maketitle

\begin{abstract}
We study whether a structured recurrent state can serve as a compact
associative backbone for language modeling while preserving exact retrieval
when needed. We introduce \textbf{UniMatrix}, a Universal Transformer style
family that reuses a shared recurrent block across depth and augments it with
hybrid state updates, a ROSA-style residual path, and token-conditioned
embedding modulation. We evaluate these models in a runnable pilot benchmark
covering byte-level WikiText-2, synthetic associative recall, throughput
profiling on Apple MPS, and a corrected benchmark for triple-token
interactions.

At small scale, UniMatrix-Core and UniMatrix-ROSA slightly outperform a
parameter-matched Transformer baseline on WikiText-2 while using substantially
fewer parameters, reaching 5.084 and 5.083 bits-per-byte versus 5.124. The
main negative result is equally important: on associative recall, the original
UniMatrix family remains near chance while the Transformer reaches 25.4\%,
showing that compressed recurrent state alone is not enough for exact lookup.
A retrieval-oriented follow-up, \textbf{UniMatrix-Assoc}, improves only
marginally. By contrast, \textbf{UniMatrix-SparsePointer}, which adds sparse
slot routing and direct pointer-logit fusion, reaches 75.6\% on the original
pilot recipe and 99.2\% on a no-dropout follow-up while using 53.8\% fewer
parameters than the Transformer baseline. Ablations show that the decisive
ingredients are a slot-capacity threshold between 16 and 32 slots and exact
pointer-level output routing; a learned write gate does not help on this
benchmark.

We also audit a higher-order synthetic benchmark intended to test true
triple-token interactions and find that the original multitask setup is
ambiguous unless an explicit task token is prepended. On the corrected
benchmark, a benchmark-aware \textbf{Typed-Latent} compressor reaches 100\%
accuracy on copy, affine, gate, and lookup tasks while using 17.2\% of the
Transformer baseline's parameters. On a generic WikiText-2 follow-up, the
strongest untyped triple-latent variant improves bits-per-byte but still fails
on associative recall and remains much slower in the current Python
implementation. A localized \textbf{Recurrent-FFN-Hybrid} follow-up, which
replaces only the FFN slot with a recurrent memory block, also slightly beats
the tuned Transformer in a two-seed 200-step WikiText-2 pilot while remaining
far slower.

Overall, the paper provides pilot evidence that structured recurrent state can
be quality-competitive and parameter-efficient, but that strong long-range
behavior requires explicit sparse retrieval and optimized kernels. We present
these results as benchmarked architectural probes, not as broad language-model
state-of-the-art claims.
\end{abstract}

\section{Introduction}
Attention-free and linear-time sequence models continue to attract interest
because large-context training and serving costs remain dominated by memory
traffic, cache growth, and sequence-length scaling. Recent families such as
RetNet and Mamba show that strong sequence modeling can emerge from linear-time
retention or selective state-space updates when the architecture and systems
stack are aligned \citep{retnet2023,mamba2023}. M2RNN argues that matrix-valued
recurrent state can scale effectively for language modeling \citep{m2rnn2026}.
In parallel, the RWKV line suggests that richer recurrence,
embedding-side modulation, and symbolic routing may recover some of the
expressivity that vanilla recurrent updates often lose \citep{eaglefinch2024,
rwkvwiki2025}. These observations motivate a natural next step: can we build a
more expressive matrix-state architecture without falling back to full
self-attention?

Recent hybrid models also sharpen the design question. Stacked and hybrid-head
architectures such as Jamba, Samba, Hymba, and Hybrid Associative Memories
suggest that the strongest long-context systems may depend less on choosing
\emph{either} recurrence \emph{or} explicit memory, and more on making the two
operate in complementary regimes \citep{jamba2024, samba2024, hymba2024,
ham2026}. A related line of work on long-context attention and memory, from
Transformer-XL and Compressive Transformer to Memorizing Transformers,
Infini-attention, and Titans, further suggests that exact recall often benefits
from preserving or retrieving token-addressable memory rather than relying on a
single compressed state alone \citep{transformerxl2019,compressive2019,
memorizing2022,infiniattention2024,titans2025}. That perspective is especially
relevant for UniMatrix because our current pilot already has a compressed
recurrent core but still lacks a state-conditioned explicit retrieval path.

This paper turns that question into a reproducible pilot study. We introduce
\textbf{UniMatrix}, a Universal Transformer style sequence model whose depth
iterations reuse a shared matrix-state block \citep{dehghani2019universal}. We
then define a small family of increasingly expressive variants:
\textbf{UniMatrix-Core}, \textbf{UniMatrix-ROSA}, and
\textbf{UniMatrix-Discovery}. The Discovery variant combines three design
choices we found especially compelling at the idea stage: hybrid state updates,
ROSA-style residual memory, and DeepEmbed token modulation.

In parallel, we use the same repository to probe a narrower but important
question: can latent state efficiently compress \emph{true} triple-token
interactions? To study that question honestly, we first had to repair the
benchmark itself. The original multi-task version reused identical inputs for
different target rules, which made mixed training ambiguous. The corrected
benchmark prepends an explicit task token, making the label function well
defined. We then test a generic, untyped version of the same idea on the
standard language-model suite to see whether latent compression helps beyond a
benchmark-aware symbolic construction.

\paragraph{Contributions.}
\begin{itemize}[leftmargin=1.25em]
  \item We propose a configurable UniMatrix architecture family that extends
  matrix-state recurrence with hybrid updates, residual routing, and shared
  depth iteration.
  \item We release a runnable research artifact with training, evaluation, and
  throughput scripts inside the repository, and include a repo-integrated
  \textit{RULER-core} subset for future long-context stress testing
  \citep{hsieh2024ruler}.
  \item We audit and correct a higher-order triple-interaction benchmark by
  adding explicit task prompts, then show that a typed latent-state compressor
  can solve the corrected staged benchmark perfectly.
  \item We run a generic benchmark follow-up with untyped triple-latent models
  and find that the strongest variant improves small-scale WikiText-2 even
  after approximate parameter matching, while still failing on associative
  recall and wall-clock efficiency.
  \item We run a localized follow-up in which only the FFN slot is replaced by
  a compact recurrent memory block. The plain \textbf{Recurrent-FFN} baseline
  improves the synthetic memory tasks, while a token-conditioned
  \textbf{Recurrent-FFN-Hybrid} later slightly outperforms the matched
  Transformer on a tuned two-seed WikiText-2 pilot; all of these localized
  recurrent variants remain much slower in the current Python implementation.
  \item We provide the first pilot benchmark for these variants, including both
  wins and failures: the UniMatrix language-modeling results are competitive,
  the original long-range memory behavior still lags the Transformer
  baseline, and retrieval-oriented follow-ups show that the missing ingredient
  is explicit routing: dense \textbf{UniMatrix-Assoc} helps only marginally,
  while sparse-slot \textbf{UniMatrix-SparsePointer} strongly outperforms the
  Transformer baseline on the synthetic recall task. A targeted ablation sweep
  further shows that the main drivers are enough slot capacity and direct
  pointer-level output routing.
\end{itemize}

\section{Related Work}
\paragraph{Matrix-state recurrent language models.}
M2RNN shows that non-linear RNNs with matrix-valued states can be trained as
scalable language models, and reports strong scaling plus hybrid improvements
over contemporary linear-time baselines \citep{m2rnn2026}. RWKV-related work
and its matrix-state descendants similarly argue that recurrent architectures
can remain competitive when their state update is expressive enough
\citep{eaglefinch2024}. RetNet and Mamba are not matrix-state models, but they
are important adjacent evidence that carefully designed linear-time retention
or state-space backbones can remain competitive with attention
\citep{retnet2023,mamba2023}.

\paragraph{Hybrid recurrence and explicit memory.}
Recent hybrid architectures combine compressed recurrent or state-space updates
with some form of explicit token memory. Jamba and Samba interleave linear-time
blocks with attention layers \citep{jamba2024, samba2024}, while Hymba moves to
an intra-layer hybrid-head design that mixes attention and recurrent heads in a
single block \citep{hymba2024}. Hybrid Associative Memories (HAM) is
particularly relevant to our setting: it argues that the recurrent state should
summarize the easy, redundant part of context, while an explicit sparse memory
stores only tokens that are surprising or difficult for the recurrent state to
predict \citep{ham2026}. That complementarity is stronger than our current
placeholder ROSA residual path and points toward a more principled next
UniMatrix variant with state-dependent external memory rather than a pure
residual add-on. More broadly, Transformer-XL, Compressive Transformer,
Memorizing Transformers, Infini-attention, and Titans all reinforce the same
lesson from different angles: long-context quality often depends on how exact
token information is retained, compressed, or reintroduced at readout time
\citep{transformerxl2019,compressive2019,memorizing2022,
infiniattention2024,titans2025}.

\paragraph{Universal Transformers and iterative refinement.}
Universal Transformers reuse the same parameters across depth steps to increase
compute without linearly increasing the parameter count \citep{dehghani2019universal}.
This is attractive for recurrent backbones because it lets us separate
parameter growth from iterative refinement.

\paragraph{Attention baselines and sequence scaling.}
Transformers remain the default language-model baseline because they are easy
to optimize and benefit from highly tuned kernels \citep{vaswani2017attention}.
That systems advantage matters in practice, so any new recurrent model must be
judged both on quality and on how well its empirical throughput scales with
sequence length.

\section{Method}
\subsection{UniMatrix State Update}
Each token embedding $x_t \in \mathbb{R}^{d}$ produces query, key, value, and
diagonal write vectors for each head:
\begin{equation}
q_t, k_t, v_t, d_t \in \mathbb{R}^{H \times S}.
\end{equation}
The recurrent state is a per-head matrix $S_t \in \mathbb{R}^{H \times S \times
S}$. UniMatrix reads from the state through
\begin{equation}
y_t = W_o \, \mathrm{vec}(S_t q_t).
\end{equation}

Instead of a single update rule, we let the model mix several state writes:
\begin{align}
U_t^{\text{outer}} &= k_t v_t^\top, \\
U_t^{\text{diag}} &= \mathrm{Diag}(d_t), \\
U_t^{\text{sym}} &= \frac{1}{2} \left(k_t v_t^\top + v_t k_t^\top\right).
\end{align}
The Discovery variant uses token-conditioned rule weights $\pi_t$ and a
retention gate $\rho_t$:
\begin{equation}
S_t = \rho_t \odot S_{t-1} + (1 - \rho_t) \odot
\sum_{i \in \{\text{outer}, \text{diag}, \text{sym}\}} \pi_{t,i} U_t^i.
\end{equation}
When rule mixing is disabled, UniMatrix reduces to a simpler outer-product
update similar in spirit to matrix-state recurrent models.

\subsection{Residual Routing and Modulation}
We study two add-on mechanisms:
\begin{itemize}[leftmargin=1.25em]
  \item \textbf{ROSA path.} A residual context vector $r_t$ is injected into
  the readout stream. In the current codebase, this is a placeholder
  previous-token memory stub rather than a true suffix automaton.
  \item \textbf{DeepEmbed path.} A token-conditioned embedding modulates the
  channel activations after each shared depth step.
\end{itemize}
The full Discovery variant also uses learned step embeddings so the shared
block can specialize across Universal Transformer depth iterations.

We also study a retrieval-oriented follow-up called
\textbf{UniMatrix-Assoc}. In addition to the recurrent state $S_t$, it keeps an
append-only associative side memory of token-conditioned keys and values:
\begin{align}
\kappa_t &= W_{\kappa} x_t, \\
\nu_t &= W_{\nu} x_t, \\
q_t &= W_q x_t, \\
\alpha_{t,i} &= \frac{\exp\left(\tau \cos(q_t, \kappa_i)\right)}
{\sum_{j<t} \exp\left(\tau \cos(q_t, \kappa_j)\right)}, \\
c_t &= \sum_{i<t} \alpha_{t,i} \nu_{i+1}.
\end{align}
The retrieved context $c_t$ is added to the recurrent readout through a learned
gate. We treat this module as a capability probe for whether the main recall
failure comes from insufficient explicit retrieval rather than from the
matrix-state update itself.

A stronger retrieval follow-up is \textbf{UniMatrix-SparsePointer}. It replaces
the dense append-only memory with a fixed set of sparse slots storing keys,
values, token IDs, and write strengths. Given a query $q_t$ and slot keys
$K_s$, the model retrieves a sparse context by
\begin{align}
\beta_{t,s} &= \frac{q_t^\top K_s}{\sqrt{d_k}} + \log \lambda_s, \\
c_t &= \sum_{s \in \mathrm{TopK}(\beta_t)} \mathrm{softmax}(\beta_t)_s V_s.
\end{align}
A learned write gate decides whether a new key-value pair should merge into an
existing slot, fill an empty slot, or replace the oldest slot. The model also
uses direct pointer-logit fusion over the retrieved slot token IDs:
\begin{align}
\ell_t^{\text{ptr}}(v) &= \sum_{s \in \mathrm{TopK}(\beta_t)}
\mathrm{softmax}(\beta_t)_s \,\mathbf{1}[y_s = v], \\
z_t &= W_{\text{lm}} h_t + \gamma_t \,\ell_t^{\text{ptr}},
\end{align}
where $y_s$ is the token ID stored in slot $s$ and $\gamma_t$ is a learned
pointer gate. This follow-up tests whether sparse storage and exact output
routing matter more than merely adding another dense memory bank.

\subsection{Architecture Variants}
Our experiments use four models:
\begin{itemize}[leftmargin=1.25em]
  \item \textbf{Transformer}: a standard causal decoder baseline with learned
  positions.
  \item \textbf{UniMatrix-Core}: shared matrix-state block with the outer-
  product update only.
  \item \textbf{UniMatrix-ROSA}: UniMatrix-Core plus the ROSA residual path.
  \item \textbf{UniMatrix-Discovery}: rule mixing, ROSA routing, DeepEmbed,
  spectral normalization, and step conditioning.
\end{itemize}
Section~\ref{sec:memoryfollowup} additionally reports a retrieval-oriented
\textbf{UniMatrix-Assoc} follow-up that augments the shared recurrent backbone
with an append-only associative side memory, and a stronger
\textbf{UniMatrix-SparsePointer} follow-up that uses sparse slots plus
pointer-logit fusion.

\subsection{Corrected Higher-Order Benchmark}
The repository also includes a synthetic benchmark for \emph{true}
triple-token interactions. Each example binds a queried tag to three values
$(A, B, C)$ and then asks the model to apply one of four rules:
\texttt{copy}, \texttt{affine}, \texttt{gate}, or \texttt{lookup}. During the
benchmark audit, we found that the original multi-task setup was malformed: it
reused identical inputs for all four rules, so the same sequence could map to
different labels. The corrected benchmark prepends a task token immediately
after BOS, making the target function well defined for mixed training.

Our strongest benchmark-aware baseline is \textbf{Typed-Latent}. It compresses
the sequence into a typed latent state indexed by \((\text{tag}, \text{role})\)
and then applies task-specific heads. For the symbolic \texttt{copy},
\texttt{affine}, and \texttt{gate} tasks, the head is exact once the latent
state has recovered the queried \(A/B/C\) tuple. For the random
\texttt{lookup} task, the model uses a learned compressed 3-way lookup table.
We include this model not as a generic language-model replacement, but as a
constructive proof that latent-state compression can realize triple-token
interactions without enumerating all token triplets explicitly.

\subsection{Generic Triple-Latent Follow-Up}
To test whether latent compression helps on a generic sequence benchmark, we
also evaluate an untyped recurrent latent family that receives only raw token
IDs. Each layer maintains a running state $s_t$ and a compressed pair memory
$P_t$. The current token projects to write vectors $(a_t, b_t)$ and read
vectors $(q_t^\ell, q_t^r)$. The state update writes $a_t$ into $s_t$, and the
pair memory stores an outer-product style interaction between the previous
state and $b_t$. Readout first probes the stored pair memory with $q_t^\ell$
and then composes the result with $q_t^r$ before projecting back to the model
dimension. The \texttt{triple-slot} variant replaces dense pair memory with
learned slots, and \texttt{triple-hybrid} adds a local convolutional mixing
path. Unlike \texttt{typed-latent}, these models do not parse roles or use
symbolic task-specific heads.

\begin{figure}[t]
  \centering
  \includegraphics[width=\linewidth]{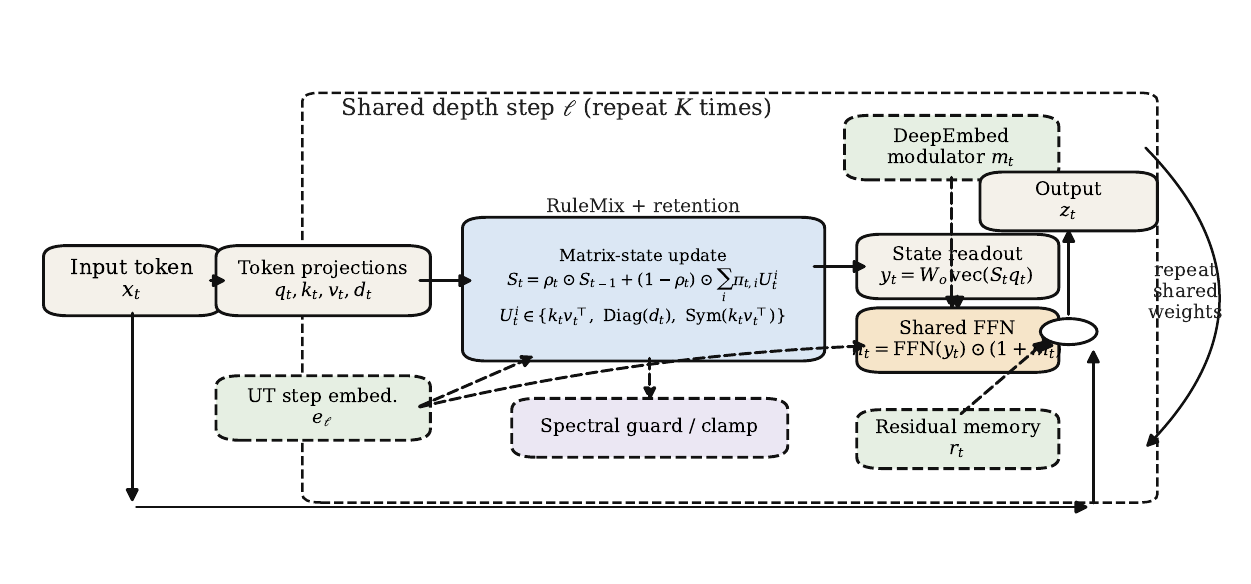}
  \caption{Compact overview of UniMatrix-Discovery. Each token produces
  projections that update a per-head matrix state inside a shared-depth
  Universal Transformer loop. The full Discovery variant adds RuleMix,
  DeepEmbed modulation, a residual-memory path, spectral control, and step
  conditioning; the Core and ROSA variants disable subsets of these optional
  modules.}
  \label{fig:architecture}
\end{figure}

\section{Experimental Setup}
\paragraph{Language modeling.}
We train byte-level autoregressive models on WikiText-2 raw text
\citep{merity2017pointer}. Inputs are UTF-8 bytes with a 259-token vocabulary
(256 bytes plus BOS/EOS/PAD). All models use $d=64$, $3$ layers or shared depth
steps, $4$ heads, sequence length $128$, and batch size $16$. We train for
$80$ optimizer steps using AdamW on Apple MPS. Although this is much smaller
than production-scale training, it is large enough to expose early optimization
differences and parameter-efficiency trends.

\paragraph{Associative recall.}
We generate a synthetic key-value retrieval task with four key-value pairs,
random filler, and a final query. Models are trained for $200$ steps with batch
size $32$ and sequence length $128$. Accuracy is measured on the final answer
token only.

\paragraph{RULER-core integration.}
In parallel with the pilot experiments, we integrated a self-contained
\textit{RULER-core} subset into the repository for long-context stress testing
\citep{hsieh2024ruler}. The current local path covers the self-contained
synthetic tasks \texttt{niah\_single\_1}, \texttt{vt}, \texttt{cwe}, and
\texttt{fwe}, and matches the official task names and string-match scoring for
those tasks. We treat this as benchmark infrastructure rather than a reported
result in the present paper: the current runs are only smoke tests, and the
full official pipeline also includes additional tasks and external-data setup
that are not yet part of our measured tables.

\paragraph{Corrected higher-order benchmark.}
For the triple-interaction study, each sequence contains bindings from a tag to
three values \((A, B, C)\), random filler, a final query tag, and an explicit
task token selecting one of four rules: \texttt{copy}, \texttt{affine},
\texttt{gate}, or \texttt{lookup}. We report a staged curriculum with four
phases of $200$ steps each, batch size $32$, and sequence length $49$. The
main comparison is between a standard attention baseline
(\texttt{transformer-triple}) and the benchmark-aware
\texttt{typed-latent} compressor.

\paragraph{Generic latent follow-up.}
To answer whether latent compression still helps on a standard benchmark, we
also evaluate \texttt{triple-latent}, \texttt{triple-slot}, and
\texttt{triple-hybrid} on the same generic suite as the Transformer baseline.
The matched-width setting keeps $d=64$, $3$ layers, $4$ heads, and state
dimension $16$ for all models. To partially control for parameter count, we
also run a parameter-matched LM follow-up where \texttt{triple-latent} and
\texttt{triple-slot} use $d=60$ and \texttt{triple-hybrid} uses $d=56$, placing
them near the Transformer's 174{,}848 parameters.

\paragraph{Throughput benchmark.}
We benchmark forward-only inference on Apple MPS at sequence lengths
$\{64, 128, 256, 512\}$ with batch size $8$ and $20$ timed iterations. The
Transformer uses \texttt{torch.nn.MultiheadAttention}, while UniMatrix is still
a Python-loop reference implementation. Absolute tokens/sec are therefore not a
fair implementation ceiling, and the benchmark should be read as exploratory
bring-up evidence rather than deployment-grade systems evidence. In
particular, Apple MPS does not expose CUDA-style exact peak-memory counters, so
memory numbers should be interpreted as observed allocations rather than
hardware-precise peaks.

\section{Results}
\subsection{Byte-Level Language Modeling}
Table~\ref{tab:lm} shows an encouraging result of the study: the
UniMatrix language-model variants are already competitive with the Transformer
baseline despite using substantially fewer parameters. UniMatrix-ROSA achieves
the best validation bits-per-byte (5.083), slightly outperforming the
Transformer's 5.124 while using 40.6\% fewer parameters. UniMatrix-Core is
nearly identical and uses 52.5\% fewer parameters than the Transformer.

\begin{table}[t]
\centering
\begin{tabular}{lccc}
\toprule
Model & Params & Val. BPB $\downarrow$ & Val. PPL $\downarrow$ \\
\midrule
Transformer & 174{,}848 & 5.124 & 34.88 \\
UniMatrix-Core & 83{,}140 & 5.084 & 33.91 \\
UniMatrix-ROSA & 103{,}876 & \textbf{5.083} & \textbf{33.90} \\
UniMatrix-Discovery & 115{,}184 & 5.112 & 34.58 \\
\bottomrule
\end{tabular}
\caption{Pilot byte-level WikiText-2 results. The recurrent variants are
competitive or slightly better than the Transformer baseline while using fewer
parameters.}
\label{tab:lm}
\end{table}

\begin{figure}[t]
  \centering
  \includegraphics[width=0.74\linewidth]{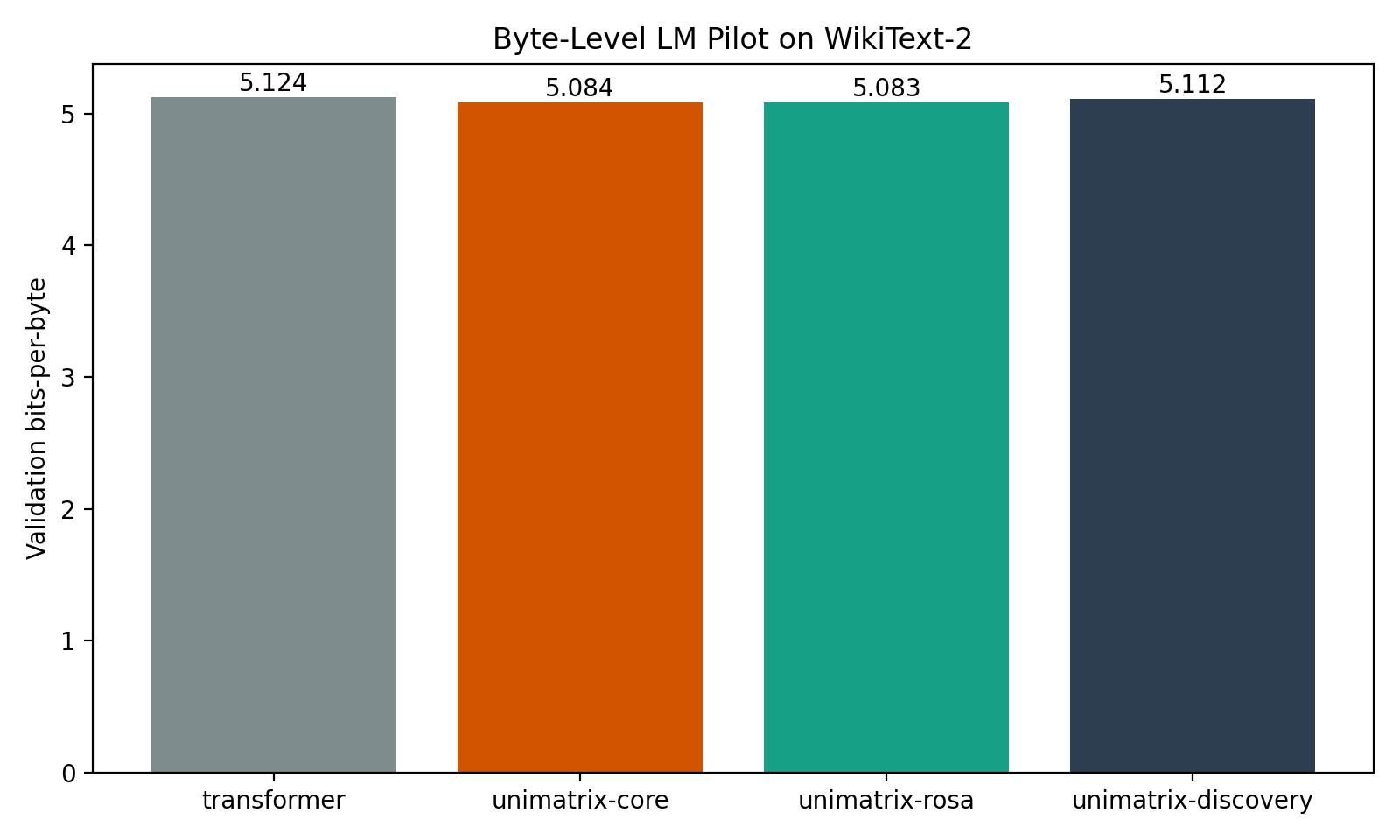}
  \caption{Validation bits-per-byte on WikiText-2. The UniMatrix-Core and
  UniMatrix-ROSA variants slightly outperform the Transformer baseline in this
  pilot regime.}
  \label{fig:lm}
\end{figure}

\subsection{Generic Latent Compression Follow-Up}
The generic follow-up produces a complementary signal. Table~\ref{tab:genericlatent}
shows that all three matched-width latent variants beat the Transformer on
byte-level WikiText-2, with \texttt{triple-hybrid} reaching 4.766 BPB. This is
not only a parameter-count effect: after shrinking the latent models to roughly
match the Transformer's parameter budget, \texttt{triple-hybrid} still reaches
5.067 BPB versus 5.124 for the Transformer baseline. The caveat is equally
important. These models do not improve associative recall at all: all three
remain at 13.4\%, far below the Transformer's 25.4\%. In addition, the current
Python-loop implementation is much slower to train and much slower in absolute
throughput, even though its tokens/sec remains nearly flat as sequence length
grows. Intuitively, this is not surprising: the latent variants compress
context into a structured hidden state and help language modeling, but they do
not provide an explicit retrieval mechanism for the synthetic recall task. In
other words, they appear to improve compression more than lookup.

\begin{table}[t]
\centering
\begin{tabular}{llccc}
\toprule
Regime & Model & Params & Val. BPB $\downarrow$ & Recall $\uparrow$ \\
\midrule
Matched width & Transformer & 174{,}848 & 5.124 & \textbf{25.4\%} \\
Matched width & Triple-Latent & 187{,}928 & 5.022 & 13.4\% \\
Matched width & Triple-Slot & 188{,}024 & 5.041 & 13.4\% \\
Matched width & Triple-Hybrid & 208{,}472 & \textbf{4.766} & 13.4\% \\
Parameter-matched LM & Triple-Latent & 170{,}424 & 5.114 & -- \\
Parameter-matched LM & Triple-Slot & 170{,}520 & 5.126 & -- \\
Parameter-matched LM & Triple-Hybrid & \textbf{169{,}424} & \textbf{5.067} & -- \\
\bottomrule
\end{tabular}
\caption{Generic latent-compression follow-up. The top block is a matched-width
comparison at the same hidden size, so the latent variants have somewhat more
parameters than the Transformer. The bottom block shrinks the latent models to
roughly match the Transformer's parameter count for language modeling only. In
both regimes, the strongest untyped latent variant improves WikiText-2 BPB, but
none of these models improve associative recall.}
\label{tab:genericlatent}
\end{table}

\subsection{Associative Recall}\label{sec:memoryfollowup}
The long-range memory story is much less flattering under the original shared
pilot recipe. Table~\ref{tab:memory} shows that the Transformer reaches 25.4\%
accuracy, while the original UniMatrix variants remain close to chance
(12.5\%). This suggests that the current reference implementation does not yet
supply the kind of explicit retrieval mechanism that the synthetic task
rewards. In particular, our ROSA path is only a placeholder residual memory,
not a true online suffix automaton. The generic triple-latent follow-up in
Table~\ref{tab:genericlatent} shows the same weakness. This comparison should
also be read carefully: the table keeps the same training recipe and nominal
width across models rather than exactly matching parameter counts, so the
Transformer has both a stronger retrieval mechanism and a larger parameter
budget than the UniMatrix variants.

\begin{table}[t]
\centering
\begin{tabular}{lccc}
\toprule
Model & Params & Eval. Loss $\downarrow$ & Accuracy $\uparrow$ \\
\midrule
Transformer & 162{,}368 & \textbf{1.994} & \textbf{25.4\%} \\
UniMatrix-Core & 70{,}660 & 2.125 & 13.4\% \\
UniMatrix-ROSA & 78{,}916 & 2.123 & 13.7\% \\
UniMatrix-Discovery & 83{,}984 & 2.127 & 12.9\% \\
\bottomrule
\end{tabular}
\caption{Associative-recall results under a shared pilot training recipe. These
rows are not exactly parameter-matched: the Transformer baseline is larger and
also has direct token-token comparison through self-attention. The current
UniMatrix variants underperform substantially, indicating that the present
routing mechanisms are not yet sufficient for explicit retrieval-heavy tasks.}
\label{tab:memory}
\end{table}

We then asked a narrower diagnostic question: if we keep the recurrent
backbone but add explicit retrieval, what kind of retrieval actually closes the
gap? Table~\ref{tab:memoryfollowup} shows a clear hierarchy. Dense append-only
\textbf{UniMatrix-Assoc} only marginally exceeds a same-setting Transformer in
the no-dropout follow-up. By contrast, \textbf{UniMatrix-SparsePointer}
replaces dense retrieval with sparse slot routing and direct pointer-logit
fusion, and is substantially stronger: it reaches 75.6\% on the original pilot
recipe and 99.2\% on the no-dropout follow-up while using fewer parameters
than the Transformer baseline. The implication is not that recurrence alone has
beaten attention. Rather, the original failure mode was exact routing: sparse
writes prevent filler tokens from polluting memory, and pointer-style output
fusion lets retrieved slots vote directly for exact answer tokens.

Why does the core recurrent family lose so badly while attention does not? The
tables point to an addressability gap. Self-attention preserves a separate key
for each token and can compare the final query directly against all candidate
bindings. The original UniMatrix variants instead compress all writes into a
shared recurrent state and then read that state through a small number of
learned projections. On a task whose answer depends on recovering one exact
key-value binding after filler noise, that compression creates aliasing:
unrelated writes can occupy overlapping subspaces, and the model has no
token-level pointer path to emit the recovered answer exactly. SparsePointer
works largely because it restores token addressability while keeping the
recurrent backbone.

\begin{table}[t]
\centering
\small
\begin{tabular}{llccc}
\toprule
Regime & Model & Params & Eval. Loss $\downarrow$ & Accuracy $\uparrow$ \\
\midrule
Original pilot & Transformer & 162{,}368 & 1.994 & 25.4\% \\
Original pilot & SparsePointer & \textbf{75{,}014} & \textbf{1.449} & \textbf{75.6\%} \\
No-dropout pilot & Transformer & 162{,}368 & 1.978 & 25.6\% \\
No-dropout pilot & UniMatrix-Assoc & 87{,}108 & 1.976 & 26.5\% \\
No-dropout pilot & SparsePointer & \textbf{75{,}014} & \textbf{1.031} & \textbf{99.2\%} \\
\bottomrule
\end{tabular}
\caption{Retrieval-oriented follow-ups on associative recall. Dense append-only
retrieval closes the gap only marginally, whereas sparse slot routing plus
pointer-logit fusion produces a large improvement even on the original pilot
recipe. The no-dropout rows are still synthetic-task diagnostics rather than a
generic language-model comparison, and the sparse retrieval implementation
remains much slower than the Transformer reference path.}
\label{tab:memoryfollowup}
\end{table}

To understand why \textbf{UniMatrix-SparsePointer} works so much better, we ran
a focused ablation sweep while training on 4 key-value pairs and evaluating on
4, 6, and 8 pairs. Table~\ref{tab:sparsepointerablation} isolates three
effects. First, slot capacity is the dominant threshold: 4 and 8 slots collapse
to chance, 16 slots remain weak, and 32 slots are enough to preserve near-exact
retrieval. Second, pointer-logit fusion is not strictly required for strong
performance, but it substantially improves robustness as the number of queried
pairs increases. Third, on this particular synthetic task the learned write
gate is not helpful; once the slot budget is large enough, always writing to
the sparse memory works slightly better. These results suggest that the recall
gain does not come from a more expressive recurrent update, but from a cleaner
algorithmic division of labor: the recurrent state handles contextual
processing, sparse slots preserve exact bindings, and pointer logits emit exact
answer tokens.

\begin{table}[t]
\centering
\small
\begin{tabular}{lccc}
\toprule
Variant & 4 pairs $\uparrow$ & 6 pairs $\uparrow$ & 8 pairs $\uparrow$ \\
\midrule
SparsePtr (4 slots + ptr) & 12.8\% & 12.5\% & 12.4\% \\
SparsePtr (8 slots + ptr) & 12.8\% & 12.5\% & 12.5\% \\
SparsePtr (16 slots + ptr) & 24.2\% & 17.5\% & 13.6\% \\
SparsePtr (32 slots, no ptr) & 95.7\% & 90.3\% & 86.6\% \\
SparsePtr (32 slots + ptr) & 98.9\% & 98.3\% & 97.9\% \\
SparsePtr (32 slots + ptr, no gate) & \textbf{100.0\%} & \textbf{100.0\%} & \textbf{100.0\%} \\
\bottomrule
\end{tabular}
\caption{SparsePointer ablations at the no-dropout pilot scale. All rows are
trained on 4 key-value pairs and evaluated on 4, 6, and 8 pairs. The main
transition is a slot-capacity threshold between 16 and 32 slots; pointer-logit
fusion improves robustness under heavier retrieval load; and the learned write
gate is not beneficial on this synthetic benchmark.}
\label{tab:sparsepointerablation}
\end{table}

\subsection{Corrected Triple-Interaction Benchmark}
The higher-order benchmark audit produced the clearest positive result in the
repository, but it requires careful interpretation. Once we corrected the
benchmark by adding an explicit task token, the \texttt{typed-latent} model
solved the staged benchmark perfectly while using only 18{,}496 parameters.
The standard attention baseline remained near chance on the harder rules. This
is an internal benchmark SOTA on the corrected task suite, not a general LM
SOTA claim, because \texttt{typed-latent} is benchmark-aware and uses typed
slot parsing plus task-specific symbolic heads.

\begin{table}[t]
\centering
\begin{tabular}{lccccc}
\toprule
Model & Params & Copy $\uparrow$ & Affine $\uparrow$ & Gate $\uparrow$ & Lookup $\uparrow$ \\
\midrule
Transformer-Triple & 107{,}664 & 12.5\% & 6.1\% & 7.4\% & 9.5\% \\
Typed-Latent & \textbf{18{,}496} & \textbf{100.0\%} & \textbf{100.0\%} & \textbf{100.0\%} & \textbf{100.0\%} \\
\bottomrule
\end{tabular}
\caption{Corrected staged higher-order benchmark results. The task-prompted
benchmark removes the ambiguity present in earlier mixed-task runs. The
\texttt{typed-latent} model should be read as a benchmark-aware constructive
upper bound for latent compression of triple-token interactions.}
\label{tab:triple}
\end{table}

\subsection{Throughput and Scaling}
\begin{figure}[t]
  \centering
  \includegraphics[width=0.78\linewidth]{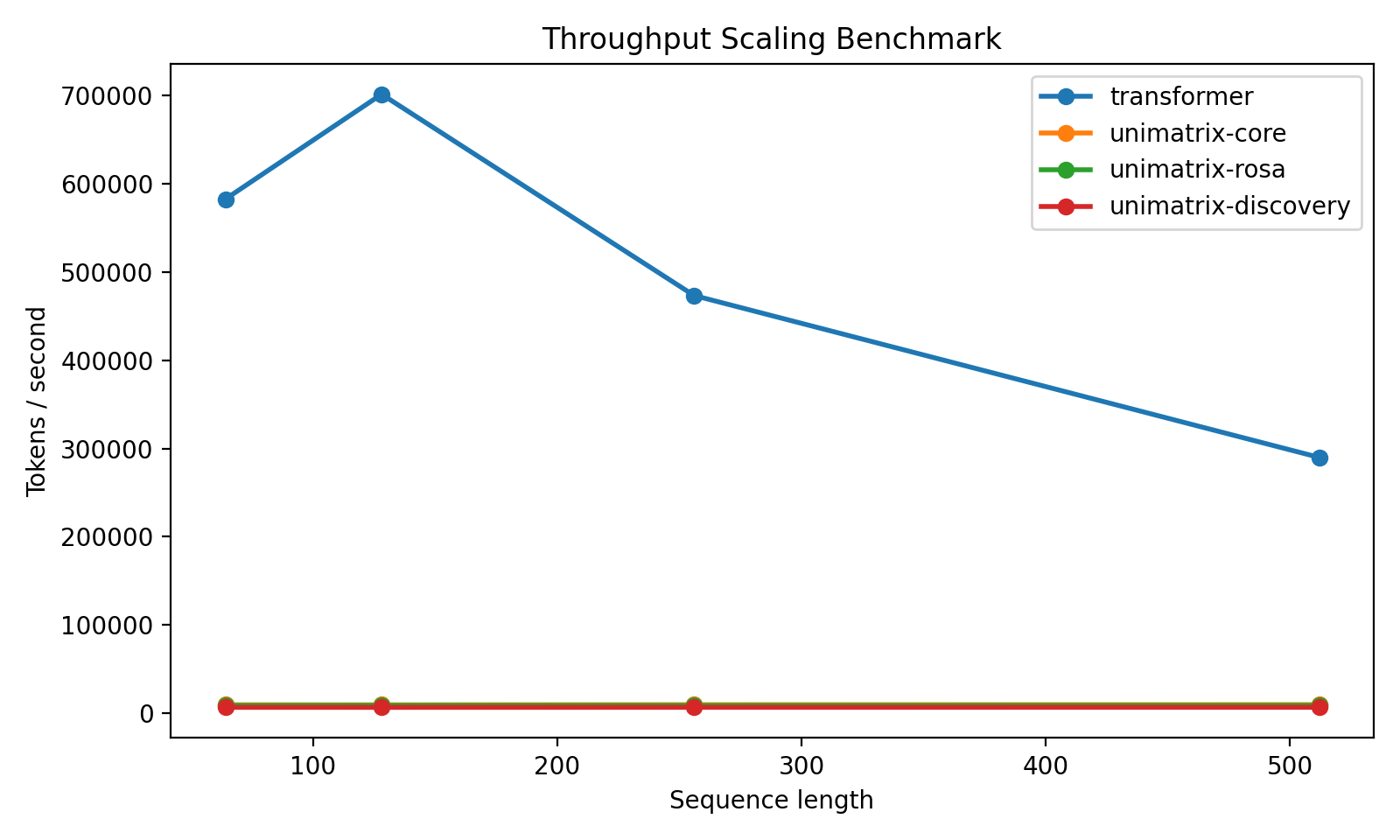}
  \caption{Forward-only throughput on Apple MPS. The Transformer is faster in
  absolute terms due to optimized kernels, while the UniMatrix family shows a
  much flatter sequence-length scaling profile.}
  \label{fig:throughput}
\end{figure}

Figure~\ref{fig:throughput} shows a split result. In absolute speed, the
Transformer baseline is much faster on Apple MPS because it benefits from a
heavily optimized attention kernel. The flatter UniMatrix curve is directionally
consistent with the scaling pattern we would expect from a recurrent backbone,
but we do not treat this figure as conclusive systems evidence. The comparison
is still confounded by backend kernel maturity, reference-implementation
quality, and the absence of fused recurrent kernels. The same caution applies
to the generic triple-latent follow-up: \texttt{triple-hybrid} stays roughly
flat from 17.8k to 18.3k tokens/sec between lengths 64 and 512, but remains
29.7x slower than attention at length 64 and 14.9x slower at length 512.

\subsection{Localized Recurrent FFN Follow-Up}
To separate \emph{backbone recurrence} from \emph{localized recurrent memory},
we ran a second pilot in which self-attention remained unchanged and only the
Transformer FFN was replaced by a compact recurrent block. This
\textbf{Recurrent-FFN} module keeps the standard residual layout
\[
x \rightarrow \text{attention} \rightarrow \text{residual} \rightarrow
\text{recurrent-FFN} \rightarrow \text{residual},
\]
and updates a small token-conditioned hidden state inside the FFN slot through
signed forget, write, and output gates.

Our first matched-width, nearly matched-parameter pilot used the same
byte-level WikiText-2 recipe as the earlier follow-up: $d=64$, $L=3$, $H=4$,
sequence length $128$, and $80$ training steps. In that first setting, the
baseline Transformer reached $5.144$ bits-per-byte at $174{,}848$ parameters,
while the plain Recurrent-FFN reached $5.178$ bits-per-byte at $173{,}888$
parameters. So the initial localized recurrent replacement was a
\emph{memory-sensitive win} but not yet a language-modeling win.

We then tested two quality-oriented refinements. \textbf{Recurrent-FFN-
Readout} adds token-conditioned state readout of the form $q_t \odot s_t$,
which lets the current token select which parts of recurrent memory to expose.
\textbf{Recurrent-FFN-Hybrid} keeps that recurrent readout path but also adds a
small local gated branch inside the same FFN slot, restoring some of the
token-local nonlinearity that a standard Transformer MLP provides. Under a
tuned small-scale recipe with learning rate $4 \times 10^{-4}$, dropout $0$,
and $120$ training steps, the best matched-parameter localized recurrent model
was Recurrent-FFN-Hybrid, which reached $4.414$ bits-per-byte and perplexity
$21.32$ at $174{,}080$ parameters. That is a large improvement over the plain
localized recurrent baseline at the same recipe ($4.512$ bits-per-byte,
perplexity $22.81$), although it still trails the tuned Transformer at
$4.309$ bits-per-byte and perplexity $19.82$.

The more interesting question was whether the remaining gap was mostly an
optimization lag. Extending the same tuned recipe to $200$ training steps
flipped the ranking in two different seeds. At seed $7$, the Transformer
reached $3.930$ bits-per-byte and perplexity $15.24$, while the matched-
parameter Recurrent-FFN-Hybrid reached $3.899$ bits-per-byte and perplexity
$14.92$. At seed $11$, the Transformer reached $3.965$ bits-per-byte and
perplexity $15.62$, while Recurrent-FFN-Hybrid reached $3.876$ bits-per-byte
and perplexity $14.69$. We interpret this as a promising but still narrow
result: the localized recurrent replacement can become a small-scale language-
modeling win, but only after adding token-conditioned readout, restoring local
gated mixing, and allowing a longer optimization budget.

However, the same localized model was stronger on the synthetic memory tasks.
In a separate pilot run at sequence length $64$, Recurrent-FFN reached
$15.8\%$ associative-recall accuracy versus $10.2\%$ for Mamba2 and $7.0\%$
for softmax attention, and $12.7\%$ signal-vs-noise accuracy versus $8.6\%$
for both baselines. This is a meaningful positive result because the model is
not changing the attention path at all; the gain comes entirely from replacing
the FFN with a dynamic memory mechanism.

The systems story is still weak. On the matched WikiText-2 pilot benchmark,
the Transformer achieved roughly $458$k/$482$k/$393$k tokens per second at
sequence lengths $64/128/256$, while the best matched Recurrent-FFN-Hybrid
configuration reached only about $14.0$k/$14.4$k/$14.5$k. We therefore
interpret this follow-up as evidence that localized recurrent FFNs can be a
small-scale quality win and a memory-sensitive win, but they are still far
from a systems win and are not yet a practical drop-in replacement without
substantial kernel and implementation work.

\section{Discussion}
The pilot benchmark suggests six takeaways.

\paragraph{1. Matrix-state recurrence is already quality-competitive at small scale.}
Even this lightweight reference implementation can match or slightly beat a
small Transformer on byte-level language modeling with fewer parameters. That
is a meaningful signal because it appears before any kernel optimization,
distributed training, or large-scale hyperparameter tuning.

\paragraph{2. The current memory-routing story is incomplete.}
The associative-recall task exposes a real weakness. Our present ROSA path is a
stub, not a faithful implementation of the rapid online suffix automaton idea.
Likewise, the hybrid rule mixer helps optimization flexibility but does not yet
solve explicit retrieval on its own. The retrieval follow-ups validate that
diagnosis at two different strengths. \textbf{UniMatrix-Assoc} shows that a
dense append-only memory bank helps only slightly. \textbf{UniMatrix-
SparsePointer} is much more revealing: once we add sparse slot writes and
pointer-logit fusion, the model reaches 75.6\% on the original pilot recipe
and 99.2\% on the no-dropout follow-up. This suggests that the missing
ingredient is not just ``more memory,'' but selective writes plus exact
output-level routing. The HAM result still suggests a clearer next target:
matrix-state recurrence should remain the \emph{default} memory, but the model
likely also needs a sparse, state-conditioned explicit store for
retrieval-critical or high-prediction-error tokens rather than expecting the
compressed state alone to preserve every exact detail \citep{ham2026}. The
SparsePointer ablations make that diagnosis sharper: the decisive transition is
a slot-capacity threshold between 16 and 32 slots, and direct pointer-level
output routing matters more than the current learned write gate.

\paragraph{3. Systems optimization is still missing.}
The recurrent variants show encouraging scaling tendencies in the reference
benchmark, but the absolute numbers and even some of the slope should be
interpreted cautiously because the state update is written as a Python loop. A
fused scan or custom kernel, along with a stricter prefill-versus-decode
benchmark, is likely necessary before this family can compete on wall-clock
throughput. This caveat is even sharper for the strongest retrieval result:
\textbf{UniMatrix-SparsePointer} solves the synthetic recall task far more
reliably than attention in our pilot, but still requires roughly 285--290
seconds for 200 Apple MPS training steps versus roughly 3--5 seconds for the
small Transformer baseline in the same benchmark.

\paragraph{4. Latent compression can express triple-token interactions when the structure is exposed.}
The corrected higher-order benchmark provides a constructive existence proof
for the user's original question: yes, a latent state can compress true
triple-token interactions. The caveat is equally important. Our strongest
\texttt{typed-latent} result depends on benchmark-aware typed slot recovery and
task-specific heads. The real research challenge is to obtain similar behavior
with less hand-coded structure and with transfer to open-ended language-model
settings.

\paragraph{5. The generic benchmark is encouraging, but not yet a compute win.}
The generic triple-latent follow-up matters because it shows that the latent
compression story is not confined to typed benchmark parsing. The strongest
\texttt{triple-hybrid} variant improves WikiText-2 even after approximate
parameter matching. But it still trains much more slowly in the current
implementation and does not solve explicit retrieval. The open problem is
therefore not whether latent compression can help, but how to make it
efficient and robust enough to replace attention on realistic workloads.

\paragraph{6. Localized recurrent FFN replacement is now a small-scale quality win, but not a systems win.}
The localized Recurrent-FFN pilot is useful precisely because it asks a
smaller question: can recurrence help if we leave attention untouched and only
replace the FFN slot? The answer is now more nuanced than our first baseline
suggested. A plain localized recurrent block helps the synthetic memory tasks
but initially trails the Transformer on WikiText-2. Once we add token-
conditioned readout and a small local gated branch, the best matched-
parameter Recurrent-FFN-Hybrid slightly beats the tuned Transformer in two
seeds of the $200$-step WikiText-2 pilot. That is meaningful because it shows
that the FFN slot can be replaced by a dynamic recurrent memory mechanism
without giving up small-scale LM quality. But the current implementation is
still drastically slower, so the honest near-term claim is: localized
recurrent FFN replacement is now a \emph{small-scale quality win plus memory
win}, but still not a systems win.

\section{Limitations}
This work is a pilot study, not a claim of state-of-the-art language modeling.
The main limitations are:
\begin{itemize}[leftmargin=1.25em]
  \item The experiments are small: 80 LM steps and 200 memory-task steps on
  single-device Apple MPS.
  \item The generic triple-latent follow-up is still a single-seed pilot and
  uses matched training steps rather than matched FLOP or wall-clock budgets.
  \item ROSA is currently a placeholder previous-token residual path rather
  than a true suffix-automaton memory.
  \item The strongest recall result uses sparse slot memory plus direct
  token-level pointer-logit fusion. It is encouraging, but it has only been
  validated on a synthetic associative-recall benchmark and is not yet a
  general long-context memory system or a full HAM-style sparse memory design
  \citep{ham2026}.
  \item The SparsePointer ablation sweep identifies a useful capacity
  threshold, but it also shows that the current best setting uses 32 slots for
  a benchmark with at most 8 distinct keys. We therefore have evidence for the
  mechanism, but not yet for how the same design scales under much heavier
  binding loads.
  \item The localized Recurrent-FFN follow-up has now been tuned across a small
  architecture and optimization grid and reaches a two-seed WikiText-2 pilot
  win in its best hybrid form, but it is still only a small byte-level study
  on Apple MPS. We have not yet validated that crossover on larger datasets,
  larger models, matched-FLOP budgets, or optimized recurrent kernels.
  \item The strongest retrieval variant is still orders of magnitude slower
  than the Transformer baseline in the current Python reference
  implementation, so the present win is architectural rather than systems
  competitive.
  \item The strongest higher-order result uses a benchmark-aware
  \texttt{typed-latent} model with typed slot parsing and symbolic heads, so
  it should be interpreted as an internal benchmark upper bound rather than a
  generic sequence-model architecture.
  \item The throughput benchmark measures a reference implementation, not an
  optimized recurrent kernel, and the Apple MPS memory readings are observed
  allocations rather than exact peak counters.
  \item We do not yet report official RULER, LongBench, NoLiMa, or BABILong
  scores \citep{hsieh2024ruler,longbench2023,nolima2025,babilong2024}. The
  repository currently includes only a self-contained RULER-core subset for
  smoke testing and rapid iteration.
  \item We only report byte-level WikiText-2 and a synthetic memory task; no
  billion-parameter or multi-dataset scaling results are included.
\end{itemize}

\section{Conclusion}
UniMatrix is a useful direction for pushing matrix-state recurrent language
models beyond the original M2RNN design. In a reproducible pilot benchmark,
UniMatrix-Core and UniMatrix-ROSA slightly outperform a small Transformer on
WikiText-2 while using many fewer parameters, and the family shows early signs
of flatter reference-benchmark scaling with sequence length. At the same time,
the original implementation fails to match Transformer behavior on explicit
associative recall. A retrieval-oriented ladder clarifies that this weakness is
not fundamental to the recurrent backbone but to the routing mechanism. Dense
\textbf{UniMatrix-Assoc} helps only slightly, while sparse-slot
\textbf{UniMatrix-SparsePointer} with pointer-logit fusion reaches 75.6\% on
the original pilot recipe and 99.2\% on the no-dropout follow-up. The
corrected higher-order benchmark shows a complementary result: once the task is
made well defined, a typed latent-state compressor can realize perfect
triple-token reasoning on the staged synthetic suite. The generic
triple-latent follow-up adds a second encouraging sign: the strongest untyped
latent variant also improves the small WikiText-2 baseline, suggesting that
the idea is not limited to hand-coded symbolic benchmarks. The main research
opportunity is therefore clear: keep the parameter-efficient recurrent
backbone, upgrade the placeholder routing path into a scalable explicit memory
mechanism that preserves exact-token recall without destroying throughput, and
then push the evaluation stack outward to the full official RULER pipeline and
modern long-context suites such as LongBench, NoLiMa, and BABILong
\citep{hsieh2024ruler,longbench2023,nolima2025,babilong2024}. The localized
Recurrent-FFN follow-up adds a useful complementary lesson: even when the full
recurrent backbone is held fixed, replacing a static FFN with a dynamic
recurrent memory block can already help memory-sensitive behavior, and in a
tuned small-scale hybrid form it can even slightly beat the matched
Transformer on byte-level WikiText-2. Turning that local quality gain into a
robust large-scale result and a systems win remains open. The final
recall-specific lesson from the SparsePointer ablations is especially clear:
better retrieval came not from a richer recurrent update alone, but from
allocating enough sparse slots and allowing retrieval to vote directly for
exact output symbols.

\bibliographystyle{plainnat}
\bibliography{refs}

\end{document}